# BayesL: a Logical Framework for the Verification of Bayesian Networks

**Stefano M. Nicoletti**
University of Twente, Enschede, the Netherlands
University of Urbino, Urbino, Italy

**E. Moritz Hahn**
University of Twente, Enschede, the Netherlands

**Mariëlle Stoelinga**
University of Twente, Enschede, the Netherlands
Radboud University, Nijmegen, the Netherlands

## Abstract

Modern explainable AI still struggles with a fundamental gap: although Bayesian networks (BNs) provide transparent probabilistic structure, there is no unified way to formally express, query, and verify what these models imply. Analysts often rely on ad hoc queries, manual interventions, or informal reasoning to explore causal relations and hypothetical scenarios, making it difficult to systematically validate model behaviour, uncover hidden assumptions, and guarantee reliability. We introduce BayesL (pronounced *Basil*), a logical framework for specifying, querying, and verifying the behaviour of BNs. BayesL is a structured language that supports both probabilistic inference queries (e.g., marginal, conditional, MAP) and model-checking–style queries that specify formal properties of BN behaviour. It facilitates versatile reasoning over causal and evidential relationships, including counterfactual what-if scenarios via conditional probability tables updates, without requiring manual modifications to the model. In addition to graph structure reasoning and inference, BayesL enables the formal specification of properties, supported by dedicated model checking algorithms and a preliminary open-source implementation. By allowing inference and verification within a single formal language, BayesL establishes a white-box verification paradigm in which model structure, assumptions, and reasoning processes are explicitly encoded and systematically checked. We demonstrate this through two diagnostic case studies and a benchmark set of BN models, showing how BayesL clarifies BN behaviour in a precise and analyzable way, advancing the transparency, trustworthiness, and practical explainability of BN-based systems.

**2012 ACM Subject Classification** Theory of computation → Logic and verification; Theory of computation → Verification by model checking

**Digital Object Identifier** 10.4230/LIPIcs...23

**Acknowledgements** This work was partially funded by the EU's Horizon 2020 Marie Curie grant No 101008233, and ERC Proof of Concept Grant 101187945 (*RUBICON*).

## 1 Introduction

Bayesian networks (BNs) are a cornerstone model in AI [25], enabling structured probabilistic reasoning under uncertainty [24, 33]. They model conditional dependencies among random variables using a directed acyclic graph (DAG), each node representing a random variable and each edge encoding a probabilistic dependency. This structure allows for the compact representation of joint probability distributions [32] enabling efficient inference and enhanced interpretability. Over the years, BNs have become integral to numerous domains [19].

**Problem statement.** Despite their ubiquitous success, a central challenge remains how to *understand* and *validate* a given BN. Especially when inferred from data [33], practitioners must ask whether a BN exhibits the properties one expects and how to explain and trace its behaviour. Addressing these questions – and guaranteeing expected behaviour – is critical for ensuring trust and reliability.

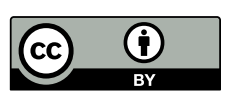

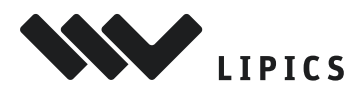

**Model checking and logics.** To this end, we leverage knowledge from the field of *model checking* [5], where given a model (a BN, in this case) and a formal specification of its expected behaviour in a logical language (our proposed logic), we can *formally guarantee* that the model exhibits that behaviour. To be comprehensive about these guarantees, users require queries that go beyond simple probability computations: structured ways to analyse how evidence propagates through the network, to verify whether key dependencies and independencies hold, and to assess the model's response to hypothetical interventions. Constructing *what-if* scenarios is an essential technique in this regard, as it allows users to evaluate the behaviour of the BN under controlled modifications and to check whether the model conforms to the expected behaviour under specific hypotheses, without the need to manually alter or reparameterize the network.

**Our contribution.** To address these needs, we propose a *Bayesian network Logic* (BayesL) – pronounced "Basil" – a framework that allows one to specify formal properties directly at the level of BNs, through expressive, layered reasoning capabilities, including marginal and conditional inference, identification of most likely outcomes, and flexible querying through non-standard comparisons. This enables practitioners to systematically probe the behaviour of the BN across multiple dimensions. Recent work highlights the need of verification approaches for BNs, notably [32] via translations to Markov models and model checking, and [1] on conditional independence in dynamic BNs. While these share a similar verification stance, they differ in focus and abstraction; and BayesL remains unique in jointly providing:

1. A language that bridges *inference* with *property specification and querying*, enabling the formulation of *what-if scenarios* without manually operating on the underlying BN model;
2. *Automatic analysis capabilities* via model checking;
3. *Formal guarantees* on the behaviour of BNs – ever so crucial in the era of AI.

**Algorithms.** Our algorithms evaluate formulae directly on BNs via a compositional, bottom-up traversal of the formula structure. The key idea is to reduce complex logical constructs to standard inference tasks by dynamically extending the BN where atoms are compiled into additional vertices. This allows us to leverage off-the-shelf inference algorithms, such as *variable elimination* and *sampling-based methods*, without requiring specialised solvers.

**Structure of the paper.** Section 2 introduces background, we then exemplify the expressive capabilities of BayesL, its syntax and semantics (Section 3), and model checking algorithms (Section 4). Section 5 introduces our implementation (Section 5.1), then validates our framework on two case studies, and on an extended benchmark set of BN models (Section 5.2). Related work and conclusions are respectively in Section 6 and Section 7.

**In summary.** We construct a logical framework to reason about BNs, for

1. Intuitive yet powerful querying, including both *inference* and *verification* queries;
2. Formulation of *what-if scenarios* within the queries;
3. *Automatic checking* of formal guarantees on BN behaviour via algorithms and an open-source *implementation* [27], for increased trust and reliability of BN reasoning.

## 2 Bayesian Networks

A BN is a tuple $B = (G, \Theta)$ where $G = (V, \to)$ is a directed acyclic graph with a finite set of vertices $V$. Each $v \in V$ represents a random variable $X_v \colon \Omega \to D$ taking values in a finite domain $D$. We assume that values in $D$ can be compared using operators $\bowtie\in\{<,\leq,=,\geq,>\}$. We often identify $v$ and $X_v$, and write $X$ for $X_v$. Each edge $v \to w$ represents the dependency of $X_v$ on $X_w$. We let $\text{parents}(v) = \{w \in V \mid w \to v\}$ be the set of incoming vertices to $v$. Each vertex $v$ is equipped with a *conditional probability table* (CPT) $\Theta_v$ that expresses the probability distribution of $X_v$ in terms of its parents: for a vertex $v$ with parents $w_1, \dots w_k$, the function $\Theta_v\colon D^k \times D \to [0,1]$ represents the conditional probability distribution $\Theta_v(d_1, \dots d_k)(d) = Pr(X_v = d \mid X_{w_1} = d_1, \dots, X_{w_k} = d_k)$. We then let $\Theta\colon \bigsqcup_{v \in V} D^{|\text{parents}(v)|} \times D \to [0,1]$ denote the CPD of the BN, where $v$ identifies the node, and the restriction $\Theta_v := \Theta|_v$ recovers the original $\Theta_v\colon D^{|\text{parents}(v)|} \times D \to [0,1]$. The semantics of a BN $B = (G, \Theta)$ is thus the joint probability function that it defines [32].

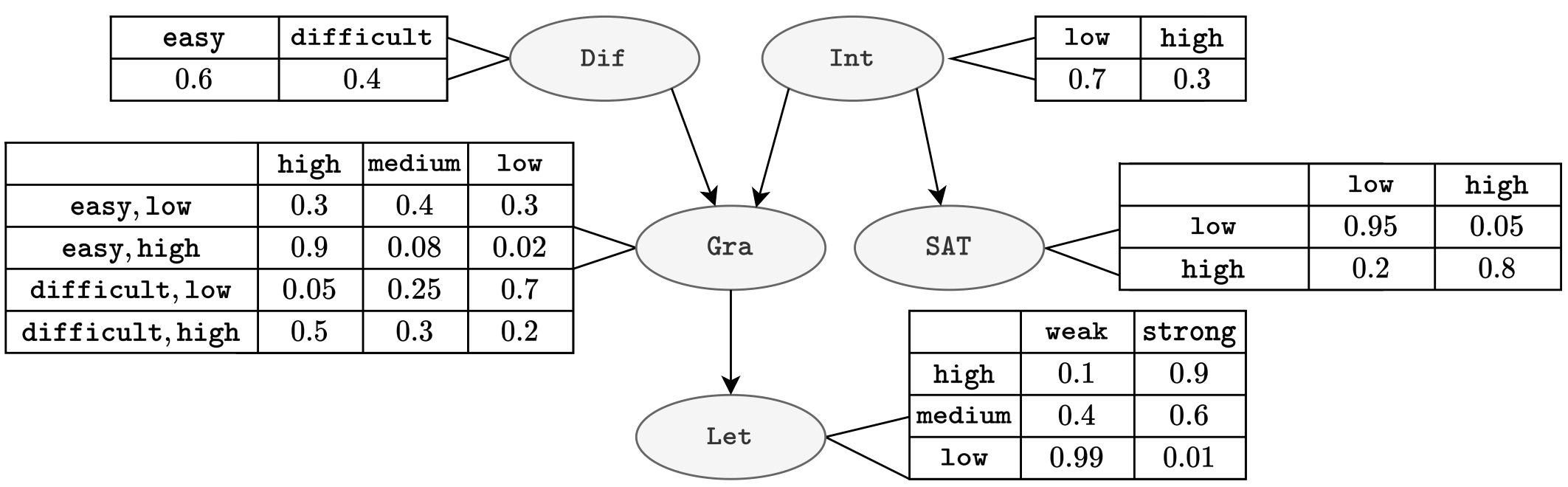

**Figure 1** BN from [19] representing dependencies between the quality of a recommendation letter, the SAT score, the grade obtained in a course, the difficulty of that course and the intelligence of a student. The image includes conditional probability tables (CPTs) for every node in the BN.

▶ **Example 1.** Figure 1 represents a well-known BN for the student example from [19]. This BN encodes dependencies between five random variables: the student's intelligence (`Int`), the course difficulty (`Dif`), the grade (`Gra`), the student's SAT score (`SAT`), and the strength of the recommendation letter they obtain (`Let`). All variables are binary-valued, except `Gra`, which is ternary-valued (i.e., *high*, *medium*, *low*). Figure 1 presents the conditional probability tables (CPTs) for each vertex. The CPT associated with a vertex $v$ specifies a probability distribution over the possible values of $v$, conditioned on a particular assignment to its parent set $\text{parents}(v)$ [32]. For instance, the CPT of *Grade* indicates that the probability of obtaining a *medium* grade (`Gra` = `medium`), given a *difficult* exam (`Dif` = `high`) and an *intelligent* student (`Int` = `high`), is 0.3.

## 2.1 Bayesian inference & reasoning

A central task in Bayesian networks (BNs) is *inference*, i.e., computing the probability of one or more random variables given observed evidence about other variables in the BN. Because a BN compactly represents a full joint distribution via its graph and conditional probability tables, inference amounts to querying this distribution in different ways. Several fundamental types of inference and reasoning are commonly distinguished (and supported by BayesL):

- **Marginal inference** computes the (unconditional) probability distribution of a variable by summing out all other variables from the joint distribution. Intuitively, it reflects the prior belief about a variable before observing any evidence. For example, P(`Let` = `strong`) represents how likely getting a *strong* letter is, regardless of other factors.
- **Conditional inference** computes the probability of a variable given observed evidence about other variables. This query type supports diagnostic reasoning (inferring causes from effects), predictive reasoning (inferring effects from causes), and intercausal reasoning [19]. E.g., $\mathrm{P}(\texttt{Let}{=}\texttt{strong} \mid \texttt{Gra}{=}\texttt{high}, \texttt{Dif}{=}\texttt{high})$ is the probability of a *strong* letter given that the grade is *high* and the exam is *difficult*.

- **MAP and MPE.** Marginal Maximum A Posteriori (MAP) identifies the most probable assignment to a subset of variables, given some evidence, while marginalizing out all remaining variables. MAP inference is useful in decision making and classification, where one seeks the most likely explanation restricted to specific variables of interest. In contrast to MAP, Most Probable Explanation (MPE) computes the most probable assignment to *all* variables in the network given the observed evidence.
- **Structural reasoning.** concerns what can be deduced directly from the graph structure of a BN, independent of specific probability values. In particular, the structure encodes *conditional independence* relations among variables [19]. This allows to determine when two variables are independent given a set of observed variables, purely from the topology. Additionally, the graph captures patterns of *influence*: variables can affect one another along active paths, while blocked paths prevent any influence. These properties clarify how dependencies arise and how they are mediated by other variables in the network.

# 3 BayesL: a Bayesian network Logic

## 3.1 BayesL Queries

Before giving the formal syntax and semantics, we illustrate BayesL via some example queries for the BN in Figure 1.

1. **What-if/CPT update** (*threshold on conditional inference with hypothesis*). After updating the probability of a *high Grade* given *high Intelligence* and a *difficult* course to 0.9, does the probability of a *strong Letter* improve beyond 0.7?

$$\mathrm{P}(\texttt{Let} = \texttt{strong} \mid \texttt{Gra} = \texttt{high}) \geq 0.7\ [\texttt{Gra} = \texttt{high} \mid\ \texttt{Int} = \texttt{high}, \texttt{Dif} = \texttt{high} \mapsto 0.9]$$

2. **MAP query** (*most probable partial explanation*). What is the most probable assignment to *Intelligence* and *SAT* given a *strong Letter*?

$$\mathrm{MAP}(\texttt{Int}, \texttt{SAT} \mid \texttt{Let} = \texttt{strong})$$

3. **MPE query** (*most probable complete explanation*). What is the most probable complete explanation for the observed *strong Letter*?

$$\mathrm{MPE}(\texttt{Let} = \texttt{strong})$$

4. **Evidential reasoning** (*diagnostic reasoning*). Given that the *Letter* is *weak*, how likely is it that the course was *difficult*?

$$\mathrm{P}(\texttt{Dif} = \texttt{high} \mid \texttt{Let} = \texttt{weak})$$

5. **Causal reasoning with disjunction.** (*predictive reasoning with disjunctive evidence*). How likely is a *strong* recommendation *Letter* due to a *high SAT* score or a *high Grade*?

$$\mathrm{P}(\texttt{Let} = \texttt{strong} \mid \texttt{SAT} = \texttt{good} \lor \texttt{Gra} = \texttt{high})$$

6. **Non-standard comparison** (*marginal reasoning with comparison and disjunction*). What is the probability that the student obtains a *Grade* lower than *high* or that the course was *easy*? (assuming an order *low* < *medium* < *high*).

$$\mathrm{P}(\texttt{Gra} < \texttt{high} \lor \texttt{Dif} = \texttt{low})$$

7. **Complex reasoning with threshold** (*threshold on predictive inference plus structural reasoning*). Is it true that the probability of a *strong Letter* given a *high Grade* is at least 0.8 and that *Intelligence* and *Letter* are independent given *Grade*?

$$\mathrm{P}(\texttt{Let}{=}\texttt{strong} \mid \texttt{Gra}{=}\texttt{high}) \geq 0.8 \wedge \mathrm{IDP}(\texttt{Int}, \texttt{Let} \mid \texttt{Gra})$$

## 3.2 BayesL Syntax

The BayesL syntax is built from atoms and further organized into three layers: Layer 1 defines probabilistic reasoning and inference – plus what-if scenarios (inline formulations of hypothetical situations w.r.t. probabilities), Layer 2 enables queries with user-specified thresholds – plus what-if scenarios, and Layer 3 captures structural reasoning.

$$\text{Atoms: } \alpha ::= X \bowtie x \mid \neg\alpha \mid \alpha \wedge \alpha \quad \text{with } \bowtie\, \in \{<, \leq, =, \geq, >\}$$

$$\text{Layer 1: } \phi ::= \mathrm{P}(\alpha \mid \alpha) \quad \phi' ::= \phi\,[X = x \mid \overline{E} = \overline{e} \mapsto q] \mid \mathrm{MAP}(\overline{X} \mid \alpha)$$

$$\text{Layer 2: } \psi ::= \phi \bowtie p \mid \psi\,[X = x \mid \overline{E} = \overline{e} \mapsto q] \mid \neg\psi \mid \psi \wedge \psi$$

$$\text{Layer 3: } \gamma ::= \mathrm{IDP}(X, Y \mid \overline{E}) \mid \neg\gamma \mid \gamma\wedge\gamma \mid \psi\wedge\gamma$$

**Syntactic sugar.** Apart from the syntax described above, we use the following minor syntactic extensions of the logic.

- Further Boolean operators such as $\vee$, $\rightarrow$, $\oplus$, $\ldots$.
- $\mathrm{MPE}(\alpha)$ defined as $\mathrm{MAP}(\overline{X} \mid \alpha)$ where $\overline{X}$ is the set of variables not occurring in $\alpha$.
- $\mathrm{INFL}(\texttt{Int}, \texttt{Let} \mid \texttt{Gra})$ – X influences Y, given E – is defined as $\neg\mathrm{IDP}(\texttt{Int}, \texttt{Let} \mid \texttt{Gra})$
- Atoms $X \bowtie Y$ where $X$ and $Y$ are variables with unique compatible variable orders, such as the domains of $X$ being `low`, `medium` and the domain of $Y$ being `medium`, `high`, leading to the joint order `low`, `medium`, `high`.

**Atoms.** At the base of BayesL are atomic *events*, i.e., formulae of the form

$$X \bowtie x$$

where $X \in V$, $x \in D$, and $\bowtie\, \in \{<, \leq, =, \geq, >\}$. Atomic formulae can be combined via Boolean connectives. We write $\overline{X} = (X_1, \ldots, X_k)$ and $\overline{x} = (x_1, \ldots, x_k)$ to denote ordered vectors of variables and corresponding values.

**Layer 1: Probabilistic inference.** Layer 1 supports (non) standard probabilistic reasoning tasks. The construct

$$\mathrm{P}(\alpha \mid \alpha)$$

denotes the conditional probability between two (Boolean) formulae in Atoms. Marginal queries are treated as a special case with omitted condition, i.e., $\mathrm{P}(\alpha)$ denotes the marginal probability of $\alpha$. This layer also supports *marginal maximum a posteriori* (MAP) – finds the most probable assignment of a subset of variables given evidence – and *most probable explanation* (MPE) inference – finds the most probable complete assignment to all variables in the network, given evidence. Specifically,

$$\mathrm{MAP}(\overline{X} \mid \alpha)$$

returns the set of assignments to $\overline{X}$ that maximize the conditional probability $\Pr(\overline{X} = \overline{x} \mid \alpha)$. To support hypothetical reasoning within queries, Layer 1 adds a local CPT update operator

$$\phi\,[X = x \mid \overline{E} = \overline{e} \mapsto q]$$

which allows evaluating how probabilities change under modifications to specific entries in the BN's conditional probability tables. This operator updates the entry for $X = x$ given $\overline{E} = \overline{e}$ to $q$, and renormalizes the remaining row values accordingly. When concatenating multiple of these operators, precedence is crucial to guarantee coherent updates on CPTs: e.g., if the same field of a CPT is updated multiple times, then the innermost update will overwrite the others.

**Layer 2: Probabilistic constraints.** Layer 2 enables reasoning about probabilistic assertions and logical combinations thereof. The main construct

$$\phi \bowtie p$$

compares the result of a Layer 1 query against a threshold $p \in [0, 1]$, using any standard comparison operator $\bowtie$. This layer also includes Boolean connectives for forming composite queries, as well as the same CPT update operator used in Layer 1, now lifted to allow evaluating the impact of updates on formulae of this layer:

$$\psi\,[X = x \mid \overline{E} = \overline{e} \mapsto q]$$

expresses the update of the $\psi$ formula via $[X = x \mid \overline{E} = \overline{e} \mapsto q]$ before its evaluation.

**Layer 3: Structural reasoning.** Layer 3 captures structural properties of the BN, independent of parameter values (i.e., only considering the graph structure $G$). Given a set of variables $\overline{E}$, we say that an *active trail* between two nodes $X$ and $Y$ exists if there is a path $\pi$ between $X$ and $Y$ in $G$ such that every triple along $\pi$ satisfies the following [19]:

- For every *chain* or *fork* structure $A \to B \to C$ or $A \leftarrow B \to C$, the node $B$ is not in $\overline{E}$,
- For every *v-structure* $A \to B \leftarrow C$, either $B \in \overline{E}$ or some descendant of $B$ is in $\overline{E}$.

If no such trail exists, then $X$ and $Y$ are *d-separated* given $\overline{E}$, i.e., *structurally conditionally independent* in all compatible distributions. The formula

$$\mathrm{IDP}(X, Y \mid \overline{E})$$

holds if $X$ and $Y$ are independent given $\overline{E}$. As before, structural formulae can be combined through Boolean connectives.

## 3.3 BayesL Semantics

**Atoms semantics.** Formulae in the Atoms layer are evaluated over full assignments $a \in D^V$, which we interpret as functions $a\colon V \to D$ assigning a value to each variable, and a BN $B$. The semantics $[\![\alpha]\!]_B(a) \in \{0, 1\}$ is defined recursively:

$$[\![X \bowtie x]\!]_B(a) = 1 \quad \text{iff } a(X) \bowtie x; \qquad [\![\neg\alpha]\!]_B(a) = 1 \quad \text{iff } [\![\alpha]\!]_B(a) = 0;$$
$$[\![\alpha_1 \wedge \alpha_2]\!]_B(a) = 1 \quad \text{iff } [\![\alpha_1]\!]_B(a) = 1 \text{ and } [\![\alpha_2]\!]_B(a) = 1$$

**Layer 1 semantics.** Formulae in Layer 1 are evaluated over a BN $B = (G, \Theta)$ and return either a probability in $[0, 1]$ or a set of assignments (for MAP and MPE queries). Let $\Pr_B(a)$ denote the joint probability of a full assignment $a \in D^V$ under $B$, computed via the standard BN factorization [19]:

$$\Pr_B(a) = \prod_{v \in V} \Theta_v\Big(a(\mathrm{parents}(v))\Big)\Big(a(v)\Big)$$

which uniquely extends to a probability measure $\Pr\colon 2^{D^V} \to [0, 1]$. Recall that $\overline{X}$ denotes a finite ordered vector of BN variables, e.g., $\overline{X} = (\texttt{Gra}, \texttt{Int}, \texttt{Dif})$, and $|\overline{X}|$ = number of variables in $\overline{X}$, e.g., $|\overline{X}| = 3$. $D^{|\overline{X}|}$ would then mean $\underbrace{D \times D \times \cdots \times D}_{|\overline{X}| \text{ times}}$ i.e., the set of all possible joint value assignments to the variables in $\overline{X}$. Then each element: $\overline{x} \in D^{|\overline{X}|}$ is a candidate joint assignment to all variables in $\overline{X}$. The semantics of Layer 1 is thus defined as follows:

$$[\![\mathrm{P}(\alpha)]\!]_B = \sum_{a \in D^V} [\![\alpha]\!]_B(a) \cdot \Pr_B(a); \quad [\![\mathrm{P}(\alpha_1 \mid \alpha_2)]\!]_B = \frac{[\![\mathrm{P}(\alpha_1 \wedge \alpha_2)]\!]_B}{[\![\mathrm{P}(\alpha_2)]\!]_B} \text{ with } [\![\mathrm{P}(\alpha_2)]\!]_B > 0;$$

$$[\![\mathrm{MAP}(\overline{X} \mid \alpha)]\!]_B = \operatorname*{argmax}_{\overline{x} \in D^{|\overline{X}|}} \Pr_B(\overline{X} = \overline{x} \mid \alpha)$$

where $\overline{V \setminus E}$ refers to an ordered vector of the variables in $V \setminus E$, with $\overline{v} \in D^{|\overline{V \setminus E}|}$ denoting corresponding assignments. Here, $\overline{E}$ denotes the vector of variables constrained in the evidence formula $\alpha$.

**CPT updates.** The update expression $\phi\,[X = x \mid \overline{E} = \overline{e} \mapsto q]$ evaluates the Layer 1 formula $\phi$ over the updated BN $B[q/(X = x \mid \overline{E} = \overline{e})] = (G, \Theta')$, where:

$$\Theta'_v = \Theta_v \quad \text{for all } v \neq X, \qquad \Theta'_X(\overline{e})(x') = \begin{cases} q & \text{if } x' = x \\ \Theta_X(\overline{e})(x') \cdot \dfrac{1-q}{1-\Theta_X(\overline{e})(x)} & \text{otherwise} \end{cases}$$

This assures that the modified row $\Theta'_X(\overline{e})$ remains a valid distribution over $D$.

**Layer 2 semantics.** Formulae in Layer 2 are evaluated over a BN $B = (G, \Theta)$ and return Boolean values. They extend Layer 1 with threshold comparisons, logical combinations, and perform CPT updates. With Booleans resolved as usual, we let semantics of $\psi$-formulae be:

$$[\![\phi \bowtie p]\!]_B = 1 \text{iff } [\![\phi]\!]_B \bowtie p; \qquad [\![\psi[X = x \mid \overline{E} = \overline{e} \mapsto q]]\!]_B = [\![\psi]\!]_{B[q/(X=x|\overline{E}=\overline{e})]}$$

**Layer 3 semantics.** Formulae in Layer 3 express properties of the BN's structure and are evaluated solely on the graph $G$: i.e., they ignore $\Theta$. With Boolean operators resolved as usual, we define: $[\![\mathrm{IDP}(X, Y \mid \overline{E})]\!]_B = 1$ iff $X$ and $Y$ are *d-separated* given $\overline{E}$ in $G$.

## 4 Model Checking

Our model checking algorithm encodes a bottom-up approach based on a recursive descent into the formula structure. During the descent, CPT updates are taken care of by modifying the BN under consideration. At the leafs of the recursive descent are the operators for conditional probability queries, MAP, and IDP. The key novelty lies in the *dynamic construction of auxiliary variables during model checking* of complex atoms $\alpha$. While augmenting BNs with auxiliary variables is an established technique – e.g., structural transformations introduce additional deterministic nodes to encode context-specific independence in [8] – such constructions are typically performed *prior to inference* and yield a fixed network. In contrast, our approach dynamically introduces new vertices during the model checking process: this allows us to use standard techniques such as *variable elimination* and *rejection sampling* to solve more expressive queries [19]. The former method works by systematically simplifying the model by removing variables until the result can be read directly. The latter is a variant of Monte Carlo simulation.

Except for IDP (which is a polynomial graph-based algorithm) these queries are NP-hard. However, using variable elimination often works well in practice, while MAP queries sometimes require significant time, see Section 5. Sampling-based methods can sometimes be faster, depending on the number of samples used, but they only provide statistical guarantees and can be problematic in the case of events with low probabilities. The runtime of the other constructs of the logic, such as Boolean operators, is negligible.

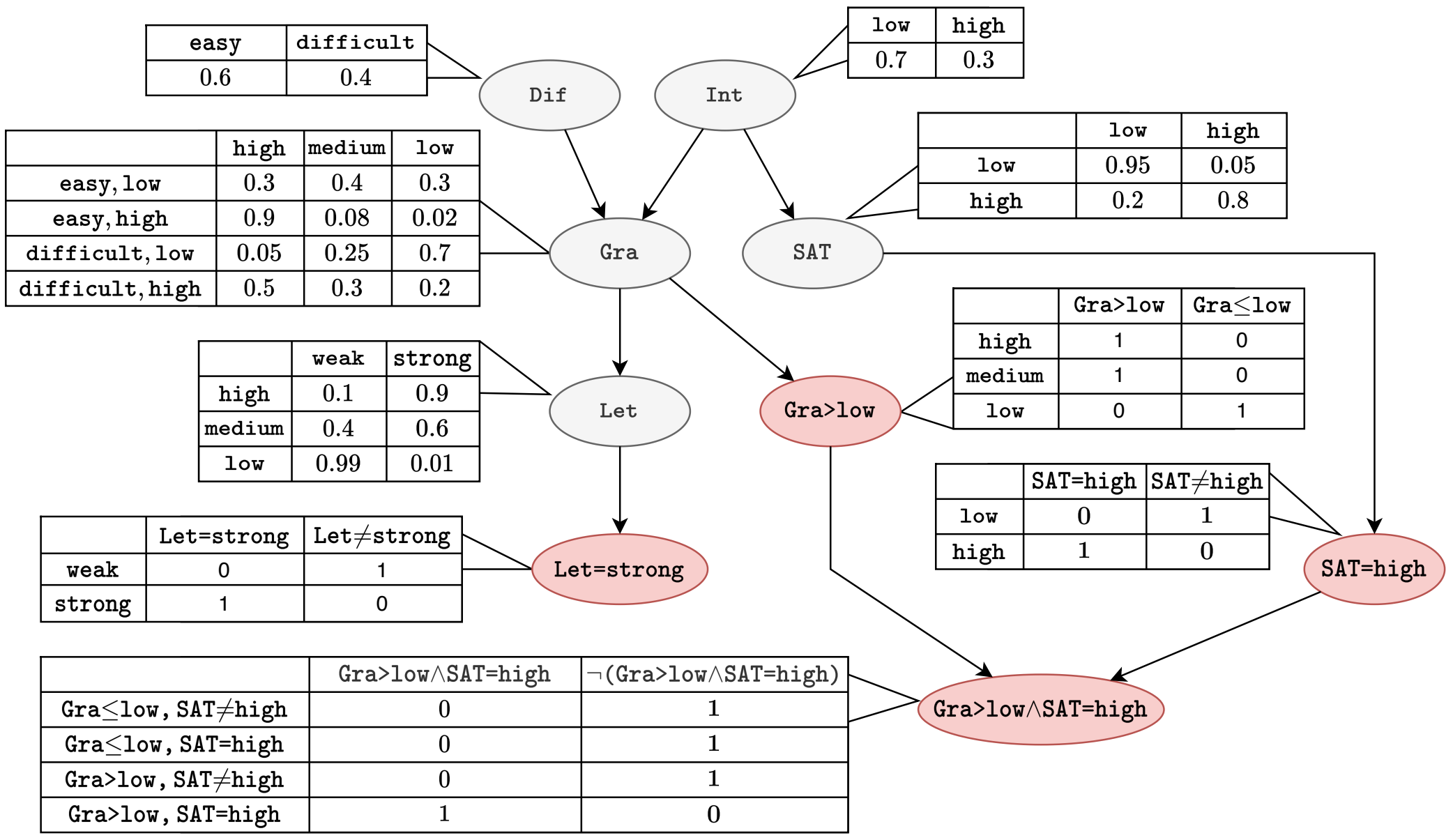

**Figure 2** Modified version of Figure 1: updated a probability assignment (red numbers) and added new vertices (in red) from atoms of an example formula.

We provide a more detailed discussion of these ideas in Algorithm 1. The starting point is calling the function MC to check whether (or with which value) formula $\chi$ holds on BN $B$. The algorithm recursively descends into subformulae of $\chi$, solving the problem in a bottom-up manner by combining results from subformulae. For the functions prefixed with COMP... (lines 2, 3, 13, and 16, we have added references to the according chapters of [19]. However, any other suitable solution method for the according problem could be employed instead. As anticipated, rather than requiring adapted versions of such algorithms for the queries of BayesL, we modify the original BN $B$. For instance, in the auxiliary function MKATMV (line 19), we create a BN which contains new vertices for the atoms used ($\dot\cup$ denotes disjoint union). If the formula contains update expressions (i.e., CPT updates/what-if scenario formulation), we employ a version of the BN in which the probabilities are changed accordingly (line 9). This way, we can take advantage of readily available implementations of these methods. Note that for compactness of the representation, the algorithm does not strictly follow the division of the logic formulae into layers. For instance, both $\gamma \wedge \gamma$ from L3 and $\psi \wedge \gamma$ from L2 are handled by line 5.

**Algorithm 1** Model Checking Algorithm MC for BayesL

**Input**: Formula $\chi$ and BN $B$; **Output**: Value according to Section 3.3

1: **function** MC($\chi$, $B$)
2: **if** $\chi = \mathrm{INFL}(X, Y \mid \overline{E})$ **then return** COMPUTEINFL$(X, Y, \overline{E})$ ▷ [19, 3.3 I. Graphs]
3: **else if** $\chi = \mathrm{IDP}(X, Y \mid \overline{E})$ **then return** COMPUTEIDP$(X, Y, \overline{E})$ ▷ [19, 3.3]
4: **else if** $\chi = \neg\chi_i$ **then return not** MC$(\chi_i, B)$
5: **else if** $\chi = \chi_l \wedge \chi_r$ **then return** MC$(\chi_l, B)$ **and** MC$(\chi_r, B)$
6: **else if** $\chi = \chi_i \bowtie p$ **then return** MC$(\chi_i, B) \bowtie p$

7: **else if** $\chi = \chi_i\,[X = x \mid \overline{E} = \overline{e} \mapsto q]$ **then**
8: $((V, \rightarrow), \Theta) := B$
9: Obtain $\Theta'$ by modifying CPT $\Theta_X$ for $X$ in $\Theta$ by update expression semantics
10: **return** $\textsc{mc}(\chi_i, ((V, \rightarrow), \Theta'))$
11: **else if** $\chi = \mathrm{P}(\alpha_l \mid \alpha_r)$ **then**
12: $B_l := \textsc{mkAtmV}(\alpha_l, B)$, $B_r := \textsc{mkAtmV}\ (\alpha_r,\ B_l)$
13: **return** $\textsc{computeConditional}(\alpha_l,\ \beta_r,\ B_r)$ ▷ [19, 12.1.3 Cond. Prob. Queries]
14: **else if** $\chi = \mathrm{MAP}(\overline{X} \mid \alpha)$ **then**
15: $B_\alpha := \textsc{mkAtmV}(\alpha, B)$
16: **return** $\textsc{compute}\mathrm{MAP}(\overline{X}, \alpha, B_\alpha)$ ▷ [19, 13 MAP Inference]
17: **end if**
18: **end function**
19: **function** $\textsc{mkAtmV}(\alpha,\ ((V, \rightarrow), \Theta))$ ▷ Handle atoms
20: **if** $\alpha = X \bowtie x$ **then**
21: $\rightarrow_\alpha\ :=\ (\rightarrow \dot{\cup} \{(X, \alpha)\})$, $\Theta^\alpha := \Theta$, $V_\alpha := V$
22: **let** $\Theta_\alpha\colon D \times \{T, F\} \to [0, 1]$ **be such that** $\Theta_\alpha(d)(T) = 1$ **if** $d \bowtie x$ **else** $0$
23: **else if** $\alpha = \neg\alpha_i$ **then**
24: $((V_\alpha, \rightarrow_i), \Theta^\alpha) := \textsc{mkAtmV}(\alpha_i, ((V, \rightarrow), \Theta))$, $\rightarrow_\alpha := (\rightarrow_i \dot{\cup} \{(\alpha_i, \alpha)\})$
25: **let** $\Theta_\alpha\colon \{T, F\} \times \{T, F\} \to [0, 1]$ **be such that** $\Theta_\alpha(F)(T) = 1$ and $\Theta_\alpha(T)(F) = 1$
26: **else if** $\alpha = \alpha_l \wedge \alpha_r$ **then**
27: $((V_l, \rightarrow_l), \Theta^l) := \textsc{mkAtmV}\ (\alpha_l,\ ((V, \rightarrow), \Theta))$
28: $((V_\alpha, \rightarrow_r), \Theta^\alpha) := \textsc{mkAtmV}(\alpha_r, ((V_l, \rightarrow_l), \Theta^l)$
29: $\rightarrow_\alpha := (\rightarrow_r \dot{\cup} \{(\alpha_l, \alpha), (\alpha_r, \alpha)\})$
30: **let** $\Theta_\alpha\colon \{T, F\}^2 \times \{T, F\} \to [0, 1]$ **be s.t.** $\Theta_\alpha(T, T)(T) = 1$ else $\Theta_\alpha(\cdot, \cdot)(F) = 1$
31: **end if**
32: **return** $((V_\alpha \dot{\cup} \{\alpha\}, \rightarrow_\alpha), \Theta^\alpha \dot{\cup} \{\Theta_\alpha\})$
33: **end function**

Correctness of the algorithm holds for the following reason: Correctness of the construction of new atoms for $\alpha$ in $\mathrm{P}(\alpha \mid \alpha)$ and $\mathrm{MAP}(\overline{X} \mid \alpha)$ holds by structural induction. Model checking of these two formula types is then correct using the assumption that the solution algorithms are correct. Correctness of nested formulae follows again by structural induction.

As an example to demonstrate how the algorithm works, consider the formula

$$\gamma = (\phi \geq 0.7)H \wedge \gamma_i \quad \text{with}$$
$$\phi = \mathrm{P}(\texttt{Let} = \texttt{strong} \mid \texttt{Gra} > \texttt{low} \wedge \texttt{SAT} = \texttt{good})$$
$$H = [\texttt{Gra} = \texttt{high} \mid \texttt{Int} = \texttt{high}, \texttt{Dif} = \texttt{high} \mapsto 0.9], \quad \gamma_i = \mathrm{IDP}(\texttt{Int}, \texttt{Let} \mid \texttt{Gra})$$

for the BN $B$ from Figure 1 (resulting in the updated BN in Figure 2). Its intuitive meaning is as follows: After updating the probability of a high grade given high intelligence and a difficult course to 0.9 ($H$), is the probability of a strong recommendation letter no smaller than 0.7, under the condition that the grade is at least medium and a good SAT score is obtained ($\phi \geq 0.7$), and are *Intelligence* and *Letter* independent given *Grade* ($\gamma_i$)?

When we apply $\textsc{mc}$ in Algorithm 1 on $\gamma$ and $B$, the algorithm executes as follows:

- Conjunction matches $(\phi \geq 0.7)H \wedge \gamma_i$ (line 5)
- Recursive descend with $(\phi \geq 0.7)H$ and $B$ (line 5)
  - Update expression matches $(\phi \geq 0.7)H$ (line 7)
  - $B$ is transformed into $\hat{B}$ with modified probabilities $\Theta_{\texttt{Grade}}(high, high) = [0.9, 0.06, 0.04]$ (line 9)

- Recursive descend with $\phi \geq 0.7$ and $\hat{B}$ (line 10)
        * Probability compare matches $\phi \geq 0.7$ (line 6)
        * Recursive descend with $\phi$ and $\hat{B}$ (line 6)
            - Conditional probability matches $\mathrm{P}(\texttt{Let} = \texttt{strong} \mid \texttt{Gra} > \texttt{low} \land \texttt{SAT} = \texttt{good})$, with $\alpha_l = (\texttt{Let} = \texttt{strong})$ and $\alpha_r = (\texttt{Gra} > \texttt{low} \land \texttt{SAT} = \texttt{good})$ (line 11)
            - Transformations applied to $\hat{B}$ using recursive function MKATMV so as to obtain $B'$ depicted in Figure 2. Subformulae $\texttt{Let} = \texttt{strong}$ and $\texttt{Gra} > \texttt{low} \land \texttt{SAT} = \texttt{good}$ have become new vertices. (line 12)
            - Because we have the two new vertices, we can now call an existing method to compute the conditional probability. We get the result $\approx 0.81$ and return it. (line 13)
        * Back at probability compare (line 6).
        * It is $0.81 \geq 0.7$. Therefore, return $T$
    - Back at update expression (line 7)
    - Return $T$.
- Back at conjunction (line 5)
- Recursive descent with $\gamma_i$ and $B$.
    - IDP matches (line 3)
    - Call existing algorithm to check whether independence holds. It does. Return $T$ (line 3)
- Back at conjunction (line 5).
- Check whether $T$ and $T$ both hold. This is indeed the case. Return $T$.

Finally, the result of MC is $T$. This means that the formula $\gamma$ holds on $B$.

## 5 Experiments

To demonstrate both the expressivity and the practical applicability of our logic, we implemented our model checking approach and evaluated it on a diverse benchmark of BNs of varying size, domain, and structural complexity from the `bnlearn` repository [34]. Across these models, we specify a rich set of properties in line with expressivity of our logic: benchmark results are summarized in Table 1.

We further present two in-depth case studies. The first is a diagnostic BN for troubleshooting printing problems in Windows 95 [15], which captures causal dependencies between system configuration, hardware components, and observed failures. The second is the MUNIN medical diagnosis network [2], a large-scale model for reasoning about neuromuscular diseases that encodes complex relationships between clinical findings and underlying pathologies. These case studies showcase how our logic can express and verify non-trivial properties in domains with very different characteristics.

For each case study, we define a representative set of properties, describe their intended meaning and interpretation of model checking result, and compare solving times of variable elimination and rejection sampling. The full set of experimental results, including additional details on all benchmark models and properties, is provided in appendix (Section A).

### 5.1 Implementation

We have implemented our method in a prototypical command-line based tool written in Lua and run by `luajit` 2.1. The method used follows the algorithm in Section 4. The solutions methods implemented for conditional probability queries and MAP are *variable elimination* and *rejection sampling*. Our tool [27] is available at `https://zenodo.org/records/19834264`.

## 5.2 Evaluation

To run our experiments, we have used an Apple M3 Pro with 18GB RAM of which we used 8GB. For the sampling-based approach, we used $10^7$ samples for all model checking runs.

**Benchmark set.** To demonstrate the expressivity of our approach, we have selected multiple models with varying numbers of variables from the model repository of the `R` library `bnlearn` [34]. Table 1 provides an overview of all models used in our experiments, including their domain, number of properties checked, size, and experimental results.

| Model | Domain | # Props | # Vars | Min (s) | Max (s) |
|---|---|---|---|---|---|
| grades | Academic grades | 9 | 5 | <1 | <1 |
| insurance | Car insurance | 25 | 27 | <1 | 4 |
| win95 | Windows 95 printer troubleshooting | 11 | 76 | <1 | 6 |
| andes | Space shuttle (ANDES) | 3 | 223 | <1 | 23 |
| pigs | Genetic analysis | 8 | 441 | <1 | 70 |
| link | Genetic analysis | 3 | 724 | <1 | 85 |
| munin | Medical diagnosis | 4 | 1041 | <1 | 152 |

**Table 1** Overview of Bayesian network models used in experiments. Minimum/maximum solving time shown for successful runs only.

Details on the full set of experiments are available in Section A. For each model, we provide specified properties, alongside: **1.** a description of the meaning of the property, **2.** the result of the model checking process (true, false, numeric value, assignment of variables to values), **3.** runtime using variable elimination, **4.** runtime using sampling, and **5.** interpretation of the result.

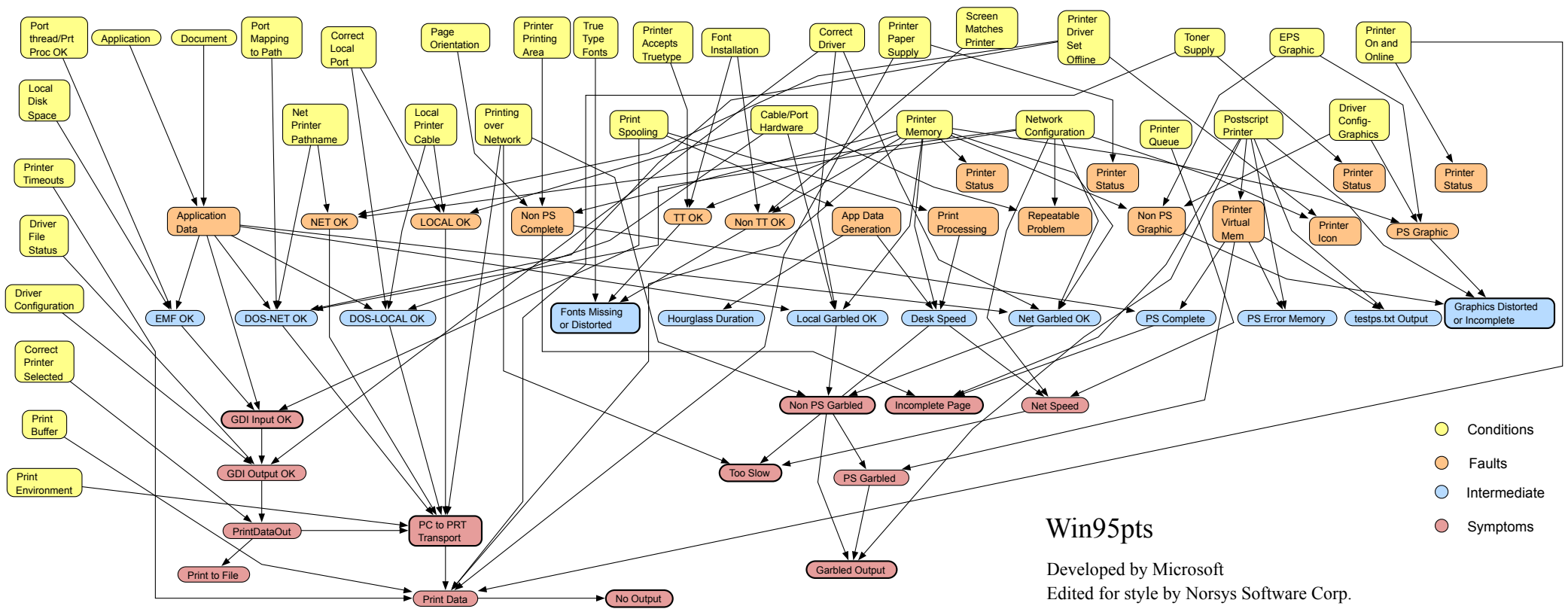

**Figure 3** BN for troubleshooting of printing problems in Windows 95. Picture taken from Norsys Software Corp. (`https://www.norsys.com/netlibrary/index.htm?net=Win95pts.htm`)

**win95: Windows 95 printer troubleshooting (76 variables).** Windows 95 printer troubleshooting diagnostic model for identifying printing problems [15]. The network models causal relationships between printer components, system settings, and observed printing failures. Variables include application status, printer drivers, network configuration, paper availability, and various system components. Following, we summarize the meaning of properties and analysis results in textual and tabular form:

- `idp`: `IDP(Problem1,NetOK|NetPrint)` **Q:** Are the problem and network status independent given the print setting? **A:** No, problem and network status are dependent given print setting
- `prob_intervention`: `P(AppOK=Correct|Problem1=No_Output∧NetPrint=Yes__Network_printer_)≥0.995[DataFile=Incorrect_Corrupt↦0.9]` **Q:** Is the probability of app correctness at least 0.995 given no output and a network printer, assuming data file corruption probability is 0.9? **A:** Yes, the probability is at least 0.995
- `map`: `MAP(NetPrint,NetOK|Problem1=No_Output)` **Q:** What are the most probable network print settings given no output? **A:** No local printer and network OK
- `prob_and`: `P(Problem1=No_Output|NetPrint=No__Local_printer_∧PrtStatToner=No_Error)<0.4` **Q:** Is the probability of no output less than 0.4 when there is no local printer and no toner error? **A:** Yes, the probability is less than 0.4
- `mpe`: `MPE(Problem1=No_Output)` **Q:** What is the most probable explanation for the printer no output problem? **A:** Multiple printer system failures
- `prob_complex`: `P(Problem1=No_Output|PrtStatToner=No_Error∧(NetPrint=No__Local_printer_∨(NetPrint=Yes__Network_printer_∧NetOK=Yes)))≤0.4` **Q:** Is the probability of no output at most 0.4 given no toner error and either no local printer or a network printer with network OK? **A:** Yes, the probability is at most 0.4
- `prob_simple`: `P(Problem1=No_Output|NetPrint=No__Local_printer_)<0.6` **Q:** Is the probability of no output less than 0.6 when the printer is not local? **A:** Yes, the probability is less than 0.6
- `prob_conjunction`: `P(Problem1=No_Output|(NetPrint=Yes__Network_printer_∧NetOK=Yes)∧PrtStatToner=No_Error)<0.4∧P(Problem1=No_Output|NetPrint=No__Local_printer_∧PrtStatToner=No_Error)<0.4` **Q:** Is the probability of no output less than 0.4 in both cases: when there is a network printer with network OK and no toner error, and when there is no local printer with no toner error? **A:** Yes, the probability is less than 0.4 in both cases

| Property | Result (var-elim) | Time (s) | |
|---|---|---|---|
| | | var-elim | sampling |
| prob_simple | `true` | <1 | <1 |
| prob_and | `true` | <1 | <1 |
| prob_complex | `true` | <1 | <1 |
| prob_conjunction | `true` | <1 | <1 |
| prob_intervention | `true` | <1 | <1 |
| map | `NetOK=Yes,NetPrint=No__Local_printer_` | <1 | 6 |
| mpe | `AppOK=Correct,...,PrtPort=Yes` | <1 | <1 |
| idp | `false` | <1 | <1 |

**munin: Medical diagnosis assistance (1041 variables).** Medical diagnostic network for neuromuscular diseases with 1041 variables representing clinical findings [2]. The network models complex probabilistic relationships between symptoms, test results, and disease states.

It assists physicians in diagnosing rare neuromuscular conditions. As done for win95, we summarize the meaning of properties and analysis results in textual and tabular form:

- `mpe`: `MPE(R_LNLW_MED_SEV=SEV)` **Q:** What is the most probable explanation for a severe medical finding? **A:** Computationally intractable for variable elimination due to model complexity
- `map`: `MAP(DIFFN_PATHO,DIFFN_TYPE|R_LNLW_MED_SEV=SEV)` **Q:** What are the most probable pathology and type given a severe finding? **A:** Specific pathology types
- `prob_cond`: `P(DIFFN_SEV=MOD|R_LNLW_MED_SEV=SEV)` **Q:** What is the probability of moderate differential severity given a severe finding? **A:** 0.5
- `prob_sev`: `P(R_LNLW_MED_SEV=SEV)` **Q:** What is the probability of a severe medical finding? **A:** 0.1

| **Property** | **Result (var-elim)** | **Time (s)** | |
|---|---|---|---|
| | | **var-elim** | **sampling** |
| prob_sev | `0.0100` | <1 | <1 |
| prob_cond | `0.5000` | <1 | <1 |
| map | `DIFFN_PATHO=AXONAL,DIFFN_TYPE=MIXED` | 152 | 122 |
| mpe | `Mem Out` | 483 | 118 |

As seen from the tables above, in most cases the time required to perform the according experiment was rather short, both using variable elimination as well as sampling. Exceptions were the MAP/MPE queries. Here, in the case the query asked to assigning values to a large number of model variables, the runtime was rather high, and in some cases our tool ran out of memory. This issue is in line with observations in [19] and hard to avoid in principle.

## 6 Related work

Several approaches have been developed to facilitate reasoning on (models similar to) BNs. We group them in:

1. Frameworks that allow/include modeling and inference on BNs: [16, 17, 4, 11, 13, 6, 22, 21, 23, 35, 9, 34];
2. Verification techniques for BNs and similar models: [32, 26, 28, 29].

### 6.1 Tools/Frameworks for Bayesian Modeling and Inference

**Semantics and calculi.** [17] offers a language for describing Bayesian phenomena in terms of programming concepts – like channel and predicate transformation – modelling inference as a calculus of string diagrams. Based on these semantics, [16] proposes a new algorithm for exact Bayesian inference. However, these works do not offer infrastructure for automatic verification of BNs' formal properties. Multi-Entity Bayesian Networks (MEBN) [21] combines BNs with first-order logic to represent complex domains involving multiple entities and relationships. While MEBN enhances the expressiveness of BNs, it adopts a proof-theoretial approach and focuses on knowledge representation, rather than on formal specification and automatic verification oriented towards model checking.

**Logic-based and relational extensions.** Bayesian Logic Programs (BLPs) combine BNs with definite clause logic, establishing a one-to-one mapping between ground atoms and random variables [18]: this allows for the representation of objects and relations, overcoming some limitations of propositional logic. However, BLPs primarily focus on the integration of logic programming with probabilistic reasoning: BayesL aims to provide verification capabilities and expressivity while remaining at the level of the BN for property specification, giving practitioners a method to query the BN model directly and avoiding lower-level querying on an intermediate translation model. Probabilistic Soft Logic (PSL) combines first-order logic with probabilistic graphical models [4] and supports efficient inference through convex optimization. However, PSL is focussed on modelling rich, structured data, esp. via the definition of hinge-loss Markov random fields and does not provide infrastructure for formal verification of BN properties. Constraint Logic Programming for Bayesian Networks (CLP(BN)) expresses BNs through the constraint logic programming framework [11]. It addresses capturing of probabilistic data using logic-based representations. CLP(BN) thus constitutes an extension of logic programming focused on defining joint probability distributions, and does not address verification of formal properties of BNs.

**Tooling for modeling, inference and analysis.** Tools like GeNIe/SMILE [13] and Bayes Server [6] provide graphical interfaces and APIs for BN modeling and inference. These tools support various functionalities, including learning, inference, and visualization. However, they lack capabilities for formal specification and verification of BN behaviour. The `bnmonitor` [22] R package allows for sensitivity and robustness analysis in BNs, and assessment of the impact of changes in the network structure or parameters. However, it does not provide a logical language for specifying and verifying formal properties. Tools like OpenBUGS [23] and Bambi [35] facilitate Bayesian analysis using (probabilistic) programming. They allow for model specification and inference but do not offer mechanisms for formal verification of BN properties. ShinyBN is an R/Shiny application that provides an interactive interface for BN modeling and inference [9]. While it enhances usability, it does not support formal specification and verification. `bnlearn` [34] is an R package for learning the graphical structure of BNs, estimating their parameters and performing probabilistic and causal inference. It provides a query language for probabilistic inference which does not cover the full expressive power of the logic we propose, as it roughly maps to L1 in our logic (see Section 3.2), with the exception of CPT updates (not present in `bnlearn`). Complex expressions are also not possible for all solution engines supported (e.g., not when using *likelihood weighting* in the method `cpquery`). Because our method relies on the creation of new vertices, we can in contrast use complex expressions for any available solution method.

## 6.2 Formal Verification Approaches

**A model checking approach.** Salmani and Katoen [32] present a method that translates BNs into tree-like Markov chains, enabling inference through reachability probabilities. This methodology aims to levereage probabilistic model checking techniques and established tooling, e.g., Storm [12] and Prism [20] model checkers. While this approach allows for formal verification, it requires the transformation of BNs into different structures and property specification is done at the level of the underlying Markov structure (with e.g., PLTL [31] and PCTL [14]). BayesL aims to provide verification capabilities and expressivity while remaining at the level of the BN for property specification: a methaporical higher-level language for BN querying, w.r.t. languages that query the underlying lower-level model translations.

**Conditional independence on dynamic BNs.** Aghamov et al. [1] study the complexity of temporal properties of conditional independence in dynamic Bayesian networks (DBNs). Thei achieve this via the use of standard LTL, augmented with atomic propositions corresponding to conditional independence statements. These statements consider both *structural* independence – via d-separation – and *stochastic* independence – with respect to the concrete probability distributions of the DBN. Their setting is orthogonal to ours: in particular, BayesL targets ordinary BNs and focuses on structural conditional independence only, without addressing temporal evolution or stochastic conditional independence*. However, it offers a higher-level, and more expressive, specification language for BN properties, closer to inference and model checking queries of practical interest, including probability comparisons, Boolean combinations, and what-if hypotheses/updates at the level of the BN itself. Hence, Aghamov et al. focus on the temporal evolution of independence properties in DBNs, while BayesL emphasizes expressive, formal specification and automatic verification of BN behaviour.

**Logics for similar risk assessment formalisms.** Recent work develops logics for reasoning on *fault trees* [26, 28] and *attack trees* [29, 30]. These contributions do not allow for reasoning on BNs, however are aligned in intent to us: i.e., they enable flexible reasoning and checking via a logic expressing queries directly on the model of interest, and not on its underlying (Markovian) representation.

## 7 Conclusion & Future Work

**Conclusion.** This contribution introduces *BayesL*, a verification framework designed to enhance trust, reliability and interpretability of BNs. BayesL supports expressive querying over BNs, including *inference*, *conditional independence* and *influence* analysis, as well as *causal* and *evidential* reasoning. A key feature is the ability to formulate *what-if* scenarios via local updates to CPT entries, without manual modification of the underlying model. The logic further supports *non-standard queries* involving threshold comparisons and flexible Boolean combinations of atomic properties, allowing users to express rich and compositional (probabilistic) queries. We further propose model checking algorithms to check BN models against property specified in the BayesL language, alongside an implementation, a case study analysis and a benchmark set.

**Future work.** Future work will focus on extending the expressivity of the logic. Firstly, we want to allow working more flexibly with probabilities, e.g., allowing queries of the form $\mathrm{P}(\ldots) - \mathrm{P}(\ldots)$ could be used to calculate the effect a certain modification of the probabilities of a BN has.

We want to improve the way in which our logic can be used to estimate *risks*. If we employ the frequently used definition that risk is impact value times impact probability, then extending our logic with constructs to describe expected values would allow it to reasoning about risks rather than just probabilities of events.

The results of MAP/MPE queries are actually stochastic events. We will allow our logic to work with such events in a more flexible way, allowing us to compute the $k$ most likely events (rather than just the most likely event), building unions of such events, and computing their probabilities.

Furthermore, we envision extensions that will leverage advances in probabilistic program synthesis [3] to support learning and constrained synthesis of BNs that satisfy formal BayesL constraints. *Constraint-guided model learning* will enable the synthesis of complete BNs that provably satisfy given logical BayesL requirements, even when starting from partially specified models.

---

*Note however that structural conditional independence implies stochastic conditional independence [1].

## A Full Experimental Results

**win95: Windows 95 printer troubleshooting (76 variables).** Windows 95 printer troubleshooting diagnostic model for identifying printing problems [15]. The network models causal relationships between printer components, system settings, and observed printing failures. Variables include application status, printer drivers, network configuration, paper availability, and various system components. Following, we summarize the meaning of properties and analysis results in textual and tabular form:

- `prob_conjunction`: `P(Problem1=No_Output|(NetPrint=Yes__Network_printer_∧NetOK=Yes)∧PrtStatToner=No_Error)<0.4∧P(Problem1=No_Output|NetPrint=No__Local_printer_∧PrtStatToner=No_Error)<0.4` **Q:** Is the probability of no output less than 0.4 in both cases: when there is a network printer with network OK and no toner error, and when there is no local printer with no toner error? **A:** Yes, the probability is less than 0.4 in both cases
- `prob_intervention`: `P(AppOK=Correct|Problem1=No_Output∧NetPrint=Yes__Network_printer_)≥0.995[DataFile=Incorrect_Corrupt↦0.9]` **Q:** Is the probability of

app correctness at least 0.995 given no output and a network printer, assuming data file corruption probability is 0.9? **A:** Yes, the probability is at least 0.995

- `map`: `MAP(NetPrint,NetOK|Problem1=No_Output)` **Q:** What are the most probable network print settings given no output? **A:** No local printer and network OK
- `map_multi`: `MAP(NetPrint,NetOK,PrtStatToner|Problem1=No_Output∧AppOK=Correct)` **Q:** What are the most probable network print settings, network status, and toner status given no output with app OK? **A:** No local printer, network OK, no toner error
- `prob_complex_cond`: `P(Problem1=No_Output|NetPrint=No__Local_printer_∧PrtStatToner=No_Error∧NetOK=Yes∧PrtPaper=Has_Paper)` **Q:** What is the probability of no output given no local printer, no toner error, network OK, and paper available? **A:** 0.3
- `prob_nested`: `P((Problem1=No_Output∧NetPrint=No__Local_printer_)∨(Problem1=No_Output∧NetOK=No))` **Q:** What is the probability of no output from either no local printer or network failure? **A:** 0.25
- `idp`: `IDP(Problem1,NetOK|NetPrint)` **Q:** Are the problem and network status independent given the print setting? **A:** No, problem and network status are dependent given print setting
- `prob_and`: `P(Problem1=No_Output|NetPrint=No__Local_printer_∧PrtStatToner=No_Error)<0.4` **Q:** Is the probability of no output less than 0.4 when there is no local printer and no toner error? **A:** Yes, the probability is less than 0.4
- `prob_simple`: `P(Problem1=No_Output|NetPrint=No__Local_printer_)<0.6` **Q:** Is the probability of no output less than 0.6 when the printer is not local? **A:** Yes, the probability is less than 0.6
- `prob_complex`: `P(Problem1=No_Output|PrtStatToner=No_Error∧(NetPrint=No__Local_printer_∨(NetPrint=Yes__Network_printer_∧NetOK=Yes)))≤0.4` **Q:** Is the probability of no output at most 0.4 given no toner error and either no local printer or a network printer with network OK? **A:** Yes, the probability is at most 0.4
- `mpe`: `MPE(Problem1=No_Output)` **Q:** What is the most probable explanation for the printer no output problem? **A:** Multiple printer system failures

| **Property** | **Result (var-elim)** | **Time (s)** | |
|---|---|---|---|
| | | **var-elim** | **sampling** |
| prob_intervention | `true` | <1 | <1 |
| map | `NetOK=Yes,NetPrint=No__Local_printer_` | <1 | 6 |
| prob_nested | `0.3743` | <1 | <1 |
| map_multi | `NetOK=Yes,PrtStatToner=No_Error,NetPrint=No__Local_printer_` | <1 | 6 |
| prob_and | `true` | <1 | <1 |
| prob_complex_cond | `0.3898` | <1 | <1 |
| prob_conjunction | `true` | <1 | <1 |
| prob_simple | `true` | <1 | <1 |
| prob_complex | `true` | <1 | <1 |
| idp | `false` | <1 | <1 |
| mpe | `AppOK=Correct,...,PrtPort=Yes` | <1 | <1 |

**munin: Medical diagnosis assistance (1041 variables).** Medical diagnostic network for neuromuscular diseases with 1041 variables representing clinical findings [2]. The network models complex probabilistic relationships between symptoms, test results, and disease states.

It assists physicians in diagnosing rare neuromuscular conditions. As done for win95, we summarize the meaning of properties and analysis results in textual and tabular form:

- `prob_sev`: `P(R_LNLW_MED_SEV=SEV)` **Q:** What is the probability of a severe medical finding? **A:** 0.1
- `map`: `MAP(DIFFN_PATHO,DIFFN_TYPE|R_LNLW_MED_SEV=SEV)` **Q:** What are the most probable pathology and type given a severe finding? **A:** Specific pathology types
- `mpe`: `MPE(R_LNLW_MED_SEV=SEV)` **Q:** What is the most probable explanation for a severe medical finding? **A:** Computationally intractable for variable elimination due to model complexity
- `prob_cond`: `P(DIFFN_SEV=MOD|R_LNLW_MED_SEV=SEV)` **Q:** What is the probability of moderate differential severity given a severe finding? **A:** 0.5

| Property | Result (var-elim) | Time (s) | |
|---|---|---|---|
| | | var-elim | sampling |
| map | `DIFFN_PATHO=AXONAL,DIFFN_TYPE=MIXED` | 152 | 122 |
| mpe | `Mem Out` | 483 | 118 |
| prob_cond | `0.5000` | <1 | <1 |
| prob_sev | `0.0100` | <1 | <1 |

**insurance: Car insurance (27 variables).** Car insurance risk assessment model covering accidents, theft, vehicle damage, and costs [7]. The network models factors affecting accident risk including driver characteristics, vehicle attributes, and environmental conditions. Variables include age, driving skill, risk aversion, vehicle make and model, safety features, and mileage.

- `map_driving`: `MAP(DrivingSkill,DrivQuality|Accident=Severe)` **Q:** What are the most probable driving skill and quality given a severe accident? **A:** SubStandard driving quality
- `prob_accident_age`: `P(Accident=Severe|Age=Senior)` **Q:** What is the probability of a severe accident given senior age? **A:** 0.15
- `prob_damage_cond`: `P(ThisCarDam=Severe|Accident=Moderate∧Antilock=True)` **Q:** What is the probability of severe damage given a moderate accident with antilock brakes? **A:** 0.45
- `prob_theft`: `P(Theft=True)` **Q:** What is the probability of car theft? **A:** 0.01
- `prob_accident_cond`: `P(Accident=Severe|Age=Senior∧Mileage=Domino)` **Q:** What is the probability of a severe accident given senior age and high mileage? **A:** 0.25
- `prob_and`: `P(Accident=Severe∧ThisCarDam=Severe)` **Q:** What is the probability of both a severe accident and severe damage? **A:** 0.02
- `mpe_theft`: `MPE(Theft=True∧HomeBase=City)` **Q:** What is the most probable explanation for theft in a city? **A:** City location and poor driving quality
- `prob_accident`: `P(Accident=Severe)` **Q:** What is the probability of a severe accident? **A:** 0.17
- `prob_ilicost`: `P(ILiCost=Million|Accident=Severe)≤0.1` **Q:** Is the probability of high liability costs at most 0.1 given a severe accident? **A:** Yes, the probability is at most 0.1
- `prob_damage`: `P(ThisCarDam=Severe|Accident=Moderate)` **Q:** What is the probability of severe car damage given a moderate accident? **A:** 0.5

- `prob_and_gt`: `P(Accident=Severe∧Age=Senior)>0.05` **Q:** Is the probability of both a severe accident and senior age greater than 0.05? **A:** No, the probability is not greater than 0.05
- `idp_multi`: `IDP(Accident,ThisCarDam|Antilock,Cushioning)` **Q:** Are accident and damage independent given safety features? **A:** No, accident and damage are dependent given safety features
- `map_risk`: `MAP(RiskAversion,MakeModel|SocioEcon=Wealthy∧Age=Adult)` **Q:** What are the most probable risk profile and car model for wealthy adults? **A:** Normal risk and luxury car
- `prob_multi_or`: `P(Accident=Mild∨Accident=Moderate∨Accident=Severe)` **Q:** What is the probability of any accident severity? **A:** 0.25
- `prob_propcost`: `P(PropCost=Thousand∨PropCost=TenThou)>0.5` **Q:** Is the probability of moderate property costs greater than 0.5? **A:** Yes, the probability is greater than 0.5
- `mpe_accident`: `MPE(Accident=Severe)` **Q:** What is the most probable explanation for a severe accident? **A:** Multiple risk factors
- `prob_medcost`: `P(MedCost=Million|Accident=Severe∧Cushioning=Poor∧Age=Senior)` **Q:** What is the probability of high medical costs given a severe accident with poor cushioning and senior age? **A:** 0.3
- `prob_accident_gt`: `P(Accident=Severe)>0.1` **Q:** Is the probability of a severe accident greater than 0.1? **A:** Yes, the probability is greater than 0.1
- `prob_theft_lt`: `P(Theft=True|HomeBase=City)<0.05` **Q:** Is the probability of theft less than 0.05 given a city home base? **A:** Yes, the probability is less than 0.05
- `prob_or`: `P(ThisCarCost=Million∨MedCost=Million)` **Q:** What is the probability of high car or medical costs? **A:** 0.01
- `prob_complex_cond`: `P(Accident=Severe|Age=Senior∧Mileage=Domino∧Antilock=False∧Cushioning=Poor)` **Q:** What is the probability of a severe accident given senior age, high mileage, no antilock brakes, and poor cushioning? **A:** 0.35
- `prob_or_gt`: `P(ThisCarCost=Million∨MedCost=Million)>0.01` **Q:** Is the probability of high costs greater than 0.01? **A:** Yes, the probability is greater than 0.01
- `prob_complex_bool`: `P((Accident=Severe∧ThisCarDam=Severe)∨(Theft=True∧PropCost=Million))` **Q:** What is the probability of either a severe accident with severe damage, or theft with expensive property damage? **A:** 0.2
- `map_multi`: `MAP(Accident,Theft,ThisCarDam|Age=Senior∧SocioEcon=Wealthy)` **Q:** What are the most probable values for accident, theft, and damage given senior age and wealthy socioeconomic status? **A:** No accident, no theft, no damage
- `idp`: `IDP(Accident,Theft|HomeBase)` **Q:** Are accident and theft independent given home base? **A:** No, accident and theft are dependent given home base

| Property | Result (var-elim) | Time (s) | |
|---|---|---|---|
| | | var-elim | sampling |
| map_driving | DrivingSkill=SubStandard,DrivQuality=Poor | <1 | 4 |
| prob_damage_cond | 0.1676 | <1 | <1 |
| prob_accident_age | 0.0740 | <1 | <1 |
| prob_accident_cond | 0.1139 | <1 | <1 |
| idp | false | <1 | <1 |
| prob_and | 0.1069 | <1 | <1 |
| mpe_theft | AntiTheft=False,...,GoodStudent=False | <1 | <1 |
| prob_ilicost | true | <1 | <1 |
| map_multi | Accident=None,Theft=False,ThisCarDam=None | <1 | 4 |
| prob_and_gt | false | <1 | <1 |
| idp_multi | false | <1 | <1 |
| map_risk | RiskAversion=Normal,MakeModel=Luxury | <1 | 4 |
| prob_accident | 0.1153 | <1 | <1 |
| prob_complex_cond | 0.1261 | 3 | <1 |
| prob_damage | 0.1905 | <1 | <1 |
| mpe_accident | OtherCar=True,...,MedCost=Thousand | <1 | <1 |
| prob_medcost | 0.3000 | <1 | <1 |
| prob_theft_lt | true | <1 | <1 |
| prob_accident_gt | true | <1 | <1 |
| prob_or | 0.0169 | <1 | <1 |
| prob_propcost | true | <1 | <1 |
| prob_multi_or | 0.2841 | <1 | <1 |
| prob_or_gt | true | <1 | <1 |
| prob_complex_bool | 0.1069 | <1 | <1 |
| prob_theft | 0.0012 | <1 | <1 |

**andes: Space shuttle (ANDES) (223 variables).** Physics problem-solving tutor model for mechanics concepts using Bayesian networks [10]. The network contains many goal and skill nodes representing physics concepts and problem-solving steps in Newtonian mechanics. It enables long-term knowledge assessment and plan recognition during student problem solving.

- `mpe_goal`: `MPE(GOAL_2=true)` **Q:** What is the most probable explanation for goal achievement? **A:** Many intermediate nodes are true
- `prob_goal`: `P(GOAL_2=true)` **Q:** What is the probability of goal achievement? **A:** 0.95
- `map_nodes`: `MAP(DISPLACEM0,GRAV2|GOAL_2=true)` **Q:** What are the most probable values of displacement and gravity given goal achievement? **A:** Displacement is false and gravity is false

| Property | Result (var-elim) | Time (s) | |
|---|---|---|---|
| | | var-elim | sampling |
| mpe_goal | `SNode_131=true,...,GOAL_130 =true` | 3 | 3 |
| prob_goal | `0.9800` | <1 | <1 |
| map_nodes | `DISPLACEM0=false,GRAV2=false` | 12 | 23 |

**pigs: Pigs network (441 variables).** Genetic linkage analysis model for pig breeding with 441 variables representing genetic markers. The network models inheritance patterns and linkage disequilibrium in pig populations for genomic selection. It is used to predict breeding values and identify candidate genes for important traits.

- `prob_nested`: `P((p630400490=0∧p48124091=0)∨(p630400490=2∧p48124091=2))` **Q:** What is the probability of matching states at either 0 or 2? **A:** 0.25
- `prob_cond`: `P(p48124091=1|p630400490=1)` **Q:** What is the probability of the second variable being 1 given the first is 1? **A:** 0.5
- `idp_simple`: `IDP(p48124091,p627270088|p630400490)` **Q:** Are the two child variables independent given the parent? **A:** Yes, children are independent given parent
- `prob_simple`: `P(p630400490=0)` **Q:** What is the probability of the first variable being 0? **A:** 0.5
- `prob_complex`: `P(p627270088=1|p630400490=1∧p48124091=1)` **Q:** What is the probability of the third variable being 1 given both parents are 1? **A:** 0.5
- `map_multi`: `MAP(p630400490,p48124091,p627270088)` **Q:** What is the most probable configuration of the three variables? **A:** No unique maximum configuration found
- `map_simple`: `MAP(p48124091,p627270088|p630400490=1)` **Q:** What are the most probable child states given parent state 1? **A:** Both children are 1
- `mpe_simple`: `MPE(p630400490=1)` **Q:** What is the most probable explanation for the first variable being 1? **A:** Most variables at state 0

| Property | Result (var-elim) | Time (s) | |
|---|---|---|---|
| | | var-elim | sampling |
| idp_simple | `true` | <1 | <1 |
| map_simple | `p627270088=1,p48124091=1` | 6 | 70 |
| prob_complex | `0.5000` | <1 | <1 |
| prob_cond | `0.5000` | <1 | <1 |
| prob_nested | `0.2500` | <1 | <1 |
| prob_simple | `0.2500` | 2 | <1 |
| map_multi | `None` | <1 | <1 |
| mpe_simple | `p82292291=0,...,p50139889=0` | 4 | 3 |

**grades: Academic grades (5 variables).** Student academic performance model relating intelligence, difficulty, SAT scores, and letter grades [19]. The network captures how student intelligence and course difficulty jointly affect academic performance through grades. SAT scores are modeled as dependent on intelligence level, while letter grades derive from the final grade.

- `mpe_letter`: `MPE(Letter=Strong)` **Q:** What is the most probable explanation for receiving a strong letter? **A:** High grade and high intelligence
- `prob_intervention`: `P(Letter=Strong|Grade=High)≥0.7[Grade=High|Intelligence=High,Difficulty=Difficult↦0.9]` **Q:** What is the probability of a strong letter given a high grade with intervention? **A:** Yes, the probability is at least 0.7

- `prob_difficulty`: `P(Difficulty=Difficult|Letter=Weak)` **Q:** What is the probability of a difficult course given a weak letter? **A:** 0.544
- `prob_complex`: `P(Letter=Strong|Intelligence=High∧Difficulty=Difficult∧SAT =High)` **Q:** What is the probability of a strong letter given high intelligence, difficult course, and high SAT? **A:** 0.5
- `map_multi`: `MAP(Grade,Difficulty,SAT|Letter=Strong)` **Q:** What are the most probable grade, difficulty, and SAT values given a strong letter? **A:** High grade, easy difficulty, high SAT
- `prob_idp`: `P(Letter=Strong|Grade=High)≥0.8∧IDP(Intelligence,Letter|Grade)` **Q:** Is the probability of a strong letter at least 0.8 given a high grade, and are intelligence and letter grade independent given grade? **A:** Yes, the probability is at least 0.8 and intelligence and letter grade are independent given grade
- `prob_comp_or`: `P(Grade<High∨Difficulty=Easy)` **Q:** What is the probability of a low grade or easy difficulty? **A:** 0.926
- `map_int_sat`: `MAP(Intelligence,SAT|Letter=Strong)` **Q:** What are the most probable intelligence and SAT scores given a strong letter? **A:** Low intelligence and low SAT scores
- `prob_or`: `P(Letter=Strong|SAT=High∨Grade=High)` **Q:** What is the probability of a strong letter given a high SAT or high grade? **A:** Yes, high SAT or high grade suffices

| **Property** | **Result (var-elim)** | **Time (s)** | |
|---|---|---|---|
| | | **var-elim** | **sampling** |
| prob_intervention | `true` | <1 | <1 |
| prob_idp | `true` | <1 | <1 |
| mpe_letter | `Difficulty=Easy,...,Intelligence=High` | <1 | <1 |
| map_multi | `Grade=High,Difficulty=Easy,SAT=Low` | <1 | <1 |
| map_int_sat | `SAT=Low,Intelligence=Low` | <1 | <1 |
| prob_complex | `0.5730` | <1 | <1 |
| prob_difficulty | `0.5442` | <1 | <1 |
| prob_comp_or | `0.9260` | <1 | <1 |
| prob_or | `0.7777` | <1 | <1 |

**link: Link network (724 variables).** Large-scale genetic linkage analysis model with 724 variables for complex genetic relationships. The network extends genetic linkage analysis with more markers and complex haplotype construction capabilities. It models detection of selection signatures in breeding populations with pedigree structure.

- `prob_allele`: `P(D0_56_d_p=a)` **Q:** What is the probability of allele A at locus 56? **A:** 0.01
- `mpe_allele`: `MPE(D0_56_d_p=a)` **Q:** What is the most probable explanation for allele A? **A:** Computationally intractable for both solvers
- `map_genotype`: `MAP(N56_d_g,N56_d_m|D0_56_d_p=a)` **Q:** What are the most probable genotype values given allele A? **A:** Genotype 1_1 with maternal allele 1

| **Property** | **Result (var-elim)** | **Time (s)** | |
|---|---|---|---|
| | | **var-elim** | **sampling** |
| mpe_allele | `Mem Out` | 221 | 218 |
| map_genotype | `Mem Out` | 106 | 85 |
| prob_allele | `0.0002` | <1 | <1 |